\documentclass[10pt,twocolumn,letterpaper]{article}

\usepackage{cvpr}              
\usepackage[dvipsnames]{xcolor}

\definecolor{cvprblue}{rgb}{0.21,0.49,0.74}
\usepackage[pagebackref,breaklinks,colorlinks,citecolor=cvprblue]{hyperref}

\def\confName{CVPR}
\def\confYear{2024}

\title{Text2Immersion: Generative Immersive Scene with 3D Gaussians}

\author{Hao Ouyang\\
HKUST\\
\and
Stephen Lombardi\\
Google\\
\and
Kathryn Heal\\
Google\\
\and
Tiancheng Sun\\
Google\\
}

\begin{document}

\twocolumn[{
\renewcommand\twocolumn[1][]{#1}
\maketitle
\begin{center}
    \vspace{-30pt}
    \includegraphics[width=0.95\linewidth]{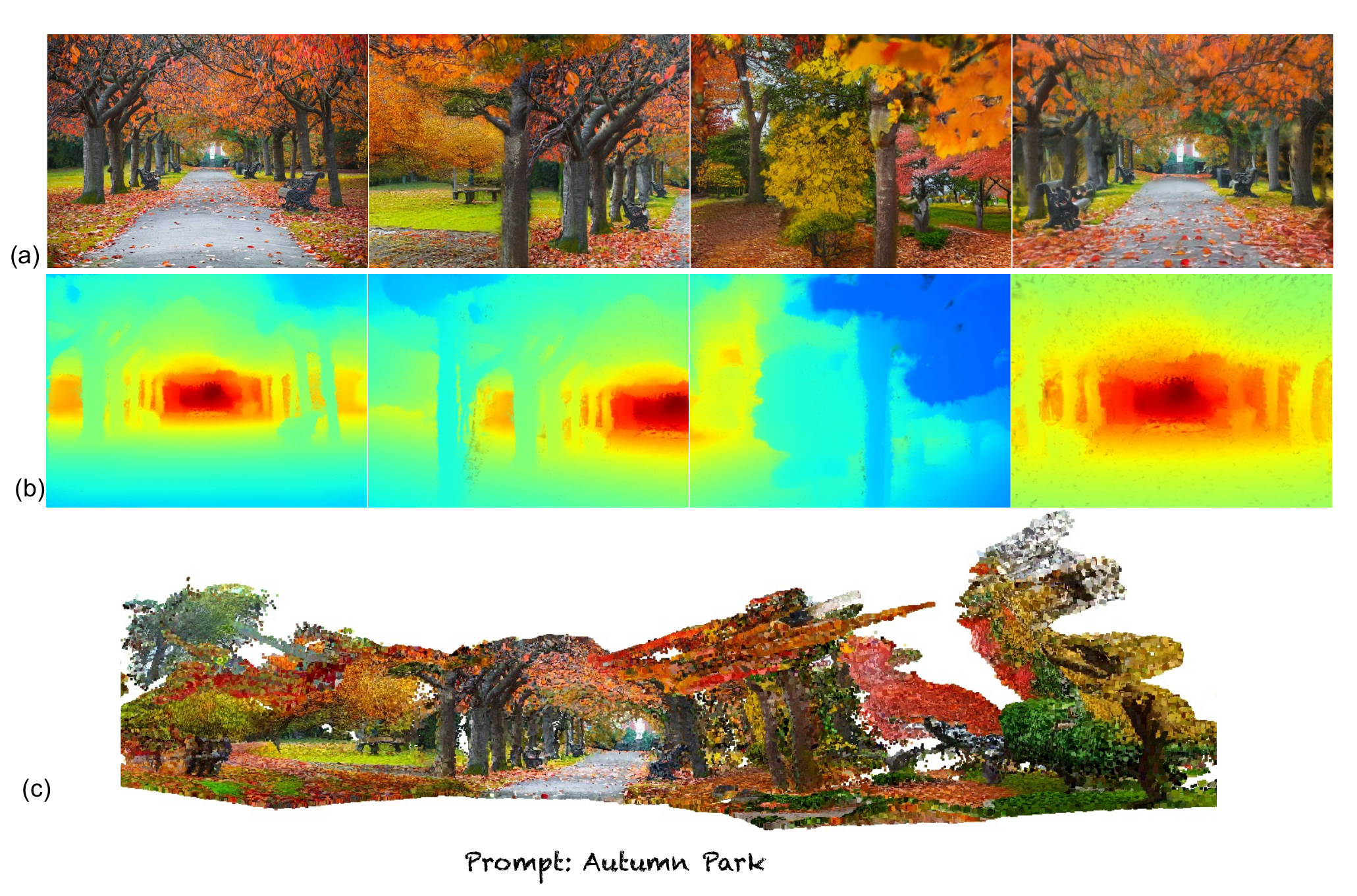}
    \vspace{-15pt}
    \captionsetup{type=figure}
    \caption{ \textbf{Generation results} of high-fidelity 3D immersive scenery of Text2Immersive pipeline. (a) Examples of renderings, as viewed from four selected directions. (b) Associated depth maps for each view, simultaneously generated by our method. (c) Point cloud demonstrating the global structure of the produced immersive scene, a collection of 3D Gaussian primitives. Shown is only the centers of the Gaussians. To capture the consistency of our generated scenes, we  encourage readers to view the accompanying supplementary videos.
    }
    \label{fig:teaser}
    \vspace{7pt}
\end{center}
}]

\maketitle
\begin{abstract}

We introduce Text2Immersion, an elegant method for producing high-quality 3D immersive scenes from text prompts. Our proposed pipeline initiates by progressively generating a Gaussian cloud using pre-trained 2D diffusion and depth estimation models. This is followed by a refining stage on the Gaussian cloud, interpolating and refining it to enhance the details of the generated scene. Distinct from prevalent methods that focus on single object or indoor scenes, or employ zoom-out trajectories, our approach generates diverse scenes with various objects, even extending to the creation of imaginary scenes. Consequently, Text2Immersion can have wide-ranging implications for various applications such as virtual reality, game development, and automated content creation. Extensive evaluations demonstrate that our system surpasses other methods in rendering quality and diversity, further progressing towards text-driven 3D scene generation. We will make the source code publicly accessible at the \href{https://ken-ouyang.github.io/text2immersion/index.html}{project page}  .
\end{abstract}    
\section{Introduction}
\label{sec:intro}

Three-dimensional content generation has numerous potential applications, most notably in gaming, film production, and virtual reality environments. The capability to generate 3D immersive environments that are rich in details can expedite the development of vibrant virtual worlds. Inspired by the recent substantial advancements in 2D generative structures~\cite{rombach2022high} and 3D representations~\cite{mildenhall2021nerf,kerbl20233d}, the field of 3D generation has similarly seen rapid progress. However, there still exists a critical gap between 2D and 3D generation. While 2D generation benefits from a wealth of training pairs, the creation of large scale 3D model datasets requires considerable human effort. Consequently, these datasets often suffer from limited diversity and realism, which limits the capability of supervised learning in 3D content generation.

To overcome the prevailing limitation of insufficient training data for 3D generation, innovative solutions with zero-shot text-to-3D generation using pretrained diffusion models have been proposed. Dreamfusion~\cite{poole2022dreamfusion} introduced the concept of Score Distillation Sampling (SDS) and distills 3D geometry and appearance from 2D diffusion models, which has significantly influenced the recent evolution of lifting methods in 3D content generation. Despite these significant advances, their generation capabilities remain predominantly constrained to 3D objects of relatively simple geometries. In contrast, an alternative approach, as employed in SceneScape~\cite{fridman2023scenescape}, Text2Room~\cite{hoellein2023text2room}, and Text2Nerf~\cite{zhang2023text2nerf}, leverages the color images generated by text-image diffusion models to guide the reconstruction of 3D scenes. However, each of these methods has its limitations. For instance, when using mesh as 3D representation, the generation process is mostly confined to indoor scenes as outdoor environments tend to distort the meshes. Similarly, when utilizing Neural Radiance Fields (NeRF)~\cite{mildenhall2021nerf} as a form of 3D representation, the methods encounter challenges with slow rendering speeds and compromised rendering quality due to insufficient supervision.

In this work, we introduce the Text2Immersion framework, a novel approach capable of generating consistently immersive, photorealistic 3D scenes while maintaining real-time rendering speeds.
Our method uses 3D Gaussians~\cite{kerbl20233d} as the 3D representations for generation, which significantly enhances rendering fidelity and improves 3D details. The proposed pipeline is composed of two stages: coarse 3D Gaussian initialization and 3D Gaussian refinement including inpainting and super-resolution.
In the initial stage, we progressively sample coarse anchor cameras and employ text-to-image diffusion models to generate or outpaint the image. Leveraging monocular depth estimation and depth alignment, we project all points to canonical coordinates, thereby constructing the initial coarse Gaussian cloud.
A pivotal challenge that arises is the construction of Gaussians from sparse views alone. Drawing inspiration from deep image priors~\cite{ulyanov2018deep}, we early stop the optimization process to maintain the rendering quality. We also observe that the training process of Gaussians, particularly with the split-and-clone operations, serves as a natural adaptive representations. This insight aids in the inpainting Gaussians using only 2D images and the direct super-resolution of Gaussians from 2D supervisions.
Through the implementation of these steps, our generated scene achieves both high-fidelity and 3D consistency. 

The pipeline proposed in this work can generate 3D photorealistic scenes that maintain consistency across varying text prompts. Each prompt yields a highly diverse set of scenes. These generated 3D scenes correspond well to the text input, as evidenced by high Clip Match scores~\cite{hessel2021clipscore}. Once a scene is generated, it can be rendered at a rate of 180 frames per second on a 3070 laptop GPU. In comparison to other scene generation methodologies, our approach attains a competitive degree of photorealism as shown in Fig.~\ref{fig:qualitative}. The generation process exhibits robust performance across 360-degree surroundings and even when handling stylized input. Our proposed framework consistently delivers superior 3D consistency and high-fidelity synthesized frames, thus validating its potential as a valuable asset in the domain of 3D scene generation.

\section{Related Work}
\label{sec:related}
\noindent\textbf{3D Representations} 
Various 3D representations have been investigated for a spectrum of 3D tasks, each posing unique advantages and challenges. Traditional 3D representations, including volume~\cite{lombardi2019neural, hologan}, point cloud~\cite{berger2014state}, multiplane images (MPI)~\cite{single_view_mpi, deng2022gram}, and meshes~\cite{munkberg21extracting}, each have distinct disadvantageous aspects when applied to 3D generation tasks. For instance, volume-based methods often entail high memory and computational costs, leading to lower resolution outcomes. Point clouds, while being flexible, fail to achieve photorealistic rendering quality, particularly when point density is insufficient. Generating meshes using neural networks presents a significant difficulty, and MPI is largely restricted to representing front-facing sceneries.

Meanwhile, implicit neural representations, especially Neural Radiance Fields (NeRF)~\cite{mildenhall2021nerf, barron2021mip}, have achieved exceptional rendering quality. However, the optimization of NeRF can be time-consuming and necessitates denser views. While various methodologies have been proposed to accelerate the optimization process, they are effective for reconstruction only. Most recently, the introduction of 3D Gaussian splatting~\cite{kerbl20233d} has made a significant impact, demonstrating remarkable quality and speed in 3D reconstruction through an efficient, differentiable renderer. Concurrent works have underscored its substantial potential in 3D object generation~\cite{tang2023dreamgaussian,yi2023gaussiandreamer}. Through this study, we further demonstrate the impressive potential of 3D Gaussians for immersive scene generation.

\noindent\textbf{Text-to-3D generation}
Generating 3D content has been a long-standing area of research focus. Initial efforts predominantly concentrated on generating 3D shapes within specific domains, such as automobiles, rooms, and faces~\cite{goodfellow2020generative,karras2019style, chan2022efficient, chan2021pi, hoellein2023text2room, niemeyer2021giraffe, liao2020towards, or2022stylesdf}. However, due to their inability to accommodate text prompts as input, these early models were limited in application. They lacked both explicit control and the capacity for high-quality texture generation. A number of other preliminary methodologies, including SynSin~\cite{wiles2020synsin} and PixelSynth~\cite{rockwell2021pixelsynth}, initiate from image inputs and directly predict the 3D geometry. As a consequence, the scenes they generate are confined to a limited viewing range around the input image.  

The rapid development of text-to-image methods~\cite{rombach2022high,dhariwal2021diffusion,ho2020denoising, song2020denoising, saharia2022photorealistic} has lead a wave of innovation in text-to-3D architectures~\cite{jun2023shap, gupta20233dgen, wang2023rodin, muller2023diffrf, watson2022novel, gu2023nerfdiff, karnewar2023holodiffusion}. However, directly applying 2D generation models to 3D content production still remains challenging, primarily due to the lack of large-scale 3D datasets. Consequently, current feed-forward text-to-3D structures are only able to generate objects within single categories. 

To achieve a broader diversity in generation results, researchers have begun to distill 3D information from 2D diffusion models. Early works in this direction, such as CLIPMesh~\cite{mohammad2022clip}, PureCLIPNeRF~\cite{lee2022understanding}, and DreamFields~\cite{Jain_2022_CVPR}, have leveraged a semantically supervised optimization strategy, under the guidance of a pretrained CLIP model~\cite{radford2021learning}, to infer shapes and textures for 3D meshes or NeRF representations. Given the extraordinary generation capabilities demonstrated by the diffusion model, recent approaches~\cite{poole2022dreamfusion, wang2023score} have sought to optimize a 3D representation to attain a high likelihood in pretrained 2D diffusion models when rendered from various viewpoints. This strategy ensures both 3D consistency and photorealism. Subsequent works~\cite{tang2023dreamgaussian, xu2023neurallift, tang2023make, lin2023magic3d, liu2023zero, liu2023one} have further enhanced training stability and generation fidelity. However, these studies have primarily concentrated on object-level generation. In contrast, our focus lies in generating immersive virtual reality scenes.

Our research shares significant parallels with recent text-to-scene studies. Works such as Text2Room~\cite{hoellein2023text2room} and SceneScape~\cite{fridman2023scenescape} employ diffusion models for the progressive generation of meshes, which, by consequence, confines the generated scenes primarily to indoor settings; any deviation from this context tends to result in considerable distortion. Conversely, Text2Nerf~\cite{zhang2023text2nerf} represents the scene with NeRF~\cite{mildenhall2021nerf}, enabling a more generalized generation. Our model adopts 3D Gaussians~\cite{kerbl20233d}, leveraging its inherent priors for multi-stage inpainting and super-resolution. This approach culminates in significantly superior rendering quality, coupled with real-time rendering speed.
 
\section{Method}
\label{sec:method}


\begin{figure*}[t]
    \centering
    \includegraphics[width=0.95\linewidth]{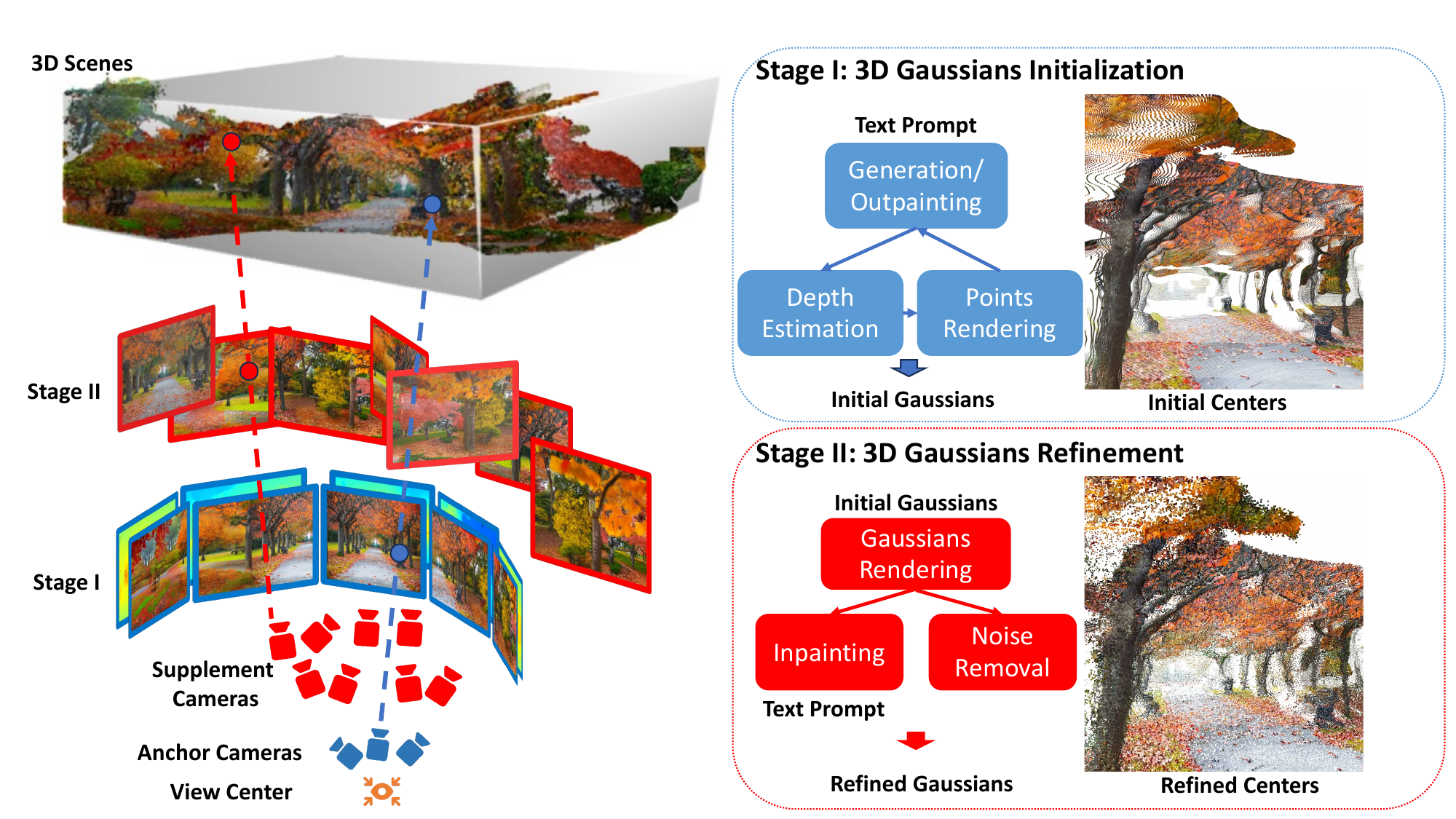}
    \caption{%
        \textbf{The overview of our pipeline.}. Our generation pipeline consists of two stages. In the first stage, we rotate the camera from the central view, and use diffusion models and monocular depth prediction modules to initialize a coarse Gaussian cloud. In the second stage, we sample more cameras around the center, and use diffusion-based inpainting modules to further refine the Gaussian cloud.
    }
    \label{fig:method}
    \vspace{-7pt}
\end{figure*}


\noindent\textbf{Problem Formulation} Given an input text prompt, our primary objective is to generate immersive 3D scenes that align with the given prompts. We choose to use 3D Gaussians~\cite{kerbl20233d} as the foundational representations, due to their rapid inference speed, superior rendering quality, and explicit nature, which can significantly benefit a broad range of applications for the generated scenes. Specifically, each Gaussian is characterized by a set of parameters, including a center $\mathbf{x} \in \mathbb R^3$, a scaling factor $\mathbf{s} \in \mathbb R^3$, a quaternion $\mathbf{q} \in \mathbb R^4$, an opacity value $\alpha \in \mathbb R$, and a base color $\mathbf{c} \in \mathbb R^3$. Consequently, our goal is to generate a set of Gaussians, as described above, which optimally represents the scene.

While 3D Gaussians have proven effective in the realm of 3D reconstruction, the task of 3D scene generation presents new challenges. As noted in the original Gaussian Splatting study~\cite{kerbl20233d}, the initial point clouds play a crucial role in the Gaussian training process. In their approach, a sparse reconstructed point cloud, acquired through COLMAP~\cite{schonberger2016structure} from multi-view input, is used for initialization. However, such initial point clouds are not readily available for generation tasks. Another significant challenge arises from the lack of consistent dense supervisions. We formulate the Gaussians from sparser views with less consistent content. The task of achieving a robust 3D representation under such limited conditions poses formidable difficulties.

To address the aforementioned challenges, we propose a pipeline consisting of two primary stages, as depicted in Fig.~\ref{fig:method}. The initial stage involves the initialization for 3D Gaussians clouds, with the specifics detailed in Section~\ref{sec:init}. We adopt a strategy that utilizes a set of anchor cameras to incrementally build the point cloud, employing depth estimation and depth alignment modules. The subsequent stage entails refining 3D Gaussians using the initial Gaussians clouds with sparse supervision. We observe interesting training priors of 3D Gaussians, leveraging these to construct superior quality inpainted and super-resoluted 3D Gaussians, as described in Section~\ref{sec:refine}.

\subsection{3D Gaussians Initialization}
\label{sec:init}
To create an immersive scene, our initial step involves generating coarse Gaussians cloud representing the scene. This construction process starts from the initial view with camera extrinsic $E_0$ and progressively expands to incorporate other $N$ anchor views, each with their respective extrinsics $\{E_1, E_2,..., E_N\}$. To simulate the experience of surrounded by immersive scenes as well as avoid the occlusions, our anchor cameras are only \textit{rotated} from the initial view. We establish two underlying assumptions for our synthetic process. First, for simplicity, $E_0$ is defined as the identity matrix. Secondly, we assume that the cameras for all synthetic views share identical intrinsics $K$. 

\noindent\textbf{Initial View Generation:} For the initial view, given the text prompt $p$, we employ the pretrained text-to-image generative model $\mathcal{F}$ to generate the synthesized image $I_0$. It's worth noting that a real image can also be utilized as the initial view. Further, we use the monocular depth estimation networks $\mathcal{F}_{depth}$ to extract the depth map $D_0$. Note that the field of view of our synthetic camera is set to the default value used in the depth estimation networks, which will reduce the estimation errors. This allows us to construct the initial points location using the formula:

\begin{align}
P_{0} = K^{-1} * E_0^{-1} * D_0,
\end{align}
where the operation $*$ applies to each pixel of $D_0$.

\noindent\textbf{Progressively Growing} Starting with the initial points, we continue the process from the nearest camera sequentially until all the anchor cameras have been traversed. For each anchor camera, we first utilize the point cloud renderer $\mathcal{R}p$ to render the existing point cloud $P_{i-1}$ for the current anchor view $i$. This process can be represented by the following equation:

\begin{align}
V_i, M_i = \mathcal{R}_p(P_{i-1}, E_i, K),
\end{align}
where $V_i$ denotes the rendered image and $M_i$ represents the corresponding visibility mask. Subsequently, we generate the regions missing from the initial render by applying the outpainting models $\mathcal{F}_{outpaint}$ with the text prompt. This process enables us to achieve the synthesized image $I_i$ for the current view:
\begin{align}
I_i = \mathcal{F}_{outpaint}(V_i, M_i).
\end{align} 

The final step involves updating the existing points with newly generated points. We employ the same depth estimation network to estimate the depth of the synthesized image $I_i$, resulting in the depth map $D_i$. This estimated depth is aligned with the existing point cloud $P_{i-1}$ by minimizing the error in the overlapping sections through a global scale and shift operation:
\begin{align}
D'_i = \text{align}(\mathcal{F}_{depth}(I_i), P{i-1}),
\end{align}
where the function align minimizes the error between the estimated depth and the existing point cloud.

The newly generated sections (i.e., areas corresponding to $1-M_i$) are then projected into the world coordinate system and integrated into the point cloud. We update point cloud by:
\begin{align}
P_i = \text{update}(P_{i-1}, D'_i, 1-M_i),
\end{align}
where the function update projects the new parts onto the world coordinates and adds them to the point cloud. Please refer to the supplement for a detailed explanation of these operations.

\noindent\textbf{Stretched Points Removal} In our approach, depth is estimated using pretrained neural models, a process which may occasionally result in common artifacts such as depth bleeding. This phenomenon typically occurs near object boundaries where there's a drastic change in depth. As a consequence, the reconstructed point cloud may contain inaccurately stretched points along these boundaries. These erroneous points can significantly impede the refining procedure. As we'll discuss in the following section, these incorrect points may generate additional errors due to the split-and-clone process. However, these stretched points are generally sparser than others. Therefore, we leverage K-nearest neighbors clustering in $P_N$ to eliminate them. Specifically, we filter out the points where the nearest distance exceeds the average distance plus two standard deviations. This method effectively prunes the initial points, ensuring a more accurate and reliable refining procedure.

With the initial point cloud, we enhance the scene's representation capability along with the rendering quality through the application of 3D Gaussians. Specifically, we utilize the point cloud to construct an initial Gaussian cloud. We align the centers of the Gaussians with the points and set the base color to match the color of the point cloud. Subsequently, the initial Gaussians $G_0$ are optimized using the sparse anchor views. Drawing from the observations made in the context of deep image priors~\cite{ulyanov2018deep}, we discovered that implementing early stopping during training is especially effective at preventing the emergence of incorrect Gaussian shapes while also achieves visually-pleasing reconstruction results. This is particularly the case when supervision isn't dense enough. As such, we limit the number of iterations during training to retrieve the optimized Gaussians $G_1$ for the initialization stage. This approach ensures a balance between training efficiency and the quality of the generated Gaussian shapes. 

\subsection{3D Gaussians Refinement}
\label{sec:refine}

As our anchor cameras are generated only by rotation but not translation, the generated scene contains a significant amount of missing regions, especially behind the generated objects in the foreground. Additionally, when the scene is zoomed in, the shape of the Gaussians becomes visible, leading to a noisy appearance in the rendering. We aim to fill in the missing regions and remove the noisy appearance during the refinement stage.

We augment the dataset by incorporating additional views, thereby ensuring robust rendering for regions occluded by foreground objects. In this stage, we sample an additional $M_1$ cameras with extrinsics of ${E_{N+1}, E_{N+2},..., E_{N+M_1}}$ around the previous anchor cameras. These additional views serve to fill in the missing regions brought about by occlusion. For these augmented views, we abstain from updating the initial point cloud. This is due to the fact that the randomness of these views could potentially lead to highly inaccurate depth estimations, such as those caused by extreme close objects. Instead, we leverage the split-and-clone process of the Gaussians to naturally cultivate new Gaussians. The split-and-clone process guarantees that the newly inpainted Gaussians appear near existing Gaussians and progressively expands outward from there (refer to the supplement for a detailed analysis). This feature enables us to inpaint 3D Gaussians using solely 2D supervision, as the initial position provides a strong prior, particularly useful for inpainting smooth surfaces. The synthesized image for a given view $i$ can be computed using the following formula: 
\begin{align}
I_{i} = \mathcal{F}_{inpaint}(\mathcal{R}(G_1, E_i, K), M_i).
\end{align}
In this equation, $\mathcal{F}_{inpaint}$ is the inpainting networks applied to the rendered Gaussians $\mathcal{R}(G_1, E_i, K)$ and the mask $M_i$ and $i$ is in $[N+1, N+M_1]$. 

Following the inpainting stage, we proceed to sample additional cameras that zoom into the scene. This strategy aids in mitigating the issue of Gaussian noise when observing the scene closely. Leveraging the split-and-clone process allows 3D Gaussians to naturally adapt for super-resolution, thereby enabling them to accommodate fine details with ease. Similar to the inpainting process, we sample $M_2$ zoomed-in views with extrinsics ${E_{N+M_1+1}, E_{N+M_1+2},..., E_{N+M_1+M_2}}$. By employing pretrained super-resolution and restoration structures, we can restore the image with a high level of detail. This augmentation further enhances the quality of the rendered scenes, ensuring more precise, detailed, and noise-free visualizations even at close range.



\noindent\textbf{Training Objectives} For optimization process in both stages, our optimization objective adheres to the same settings used in Gaussian Splatting~\cite{kerbl20233d}, incorporating both the $\mathcal{L}_1$ loss and the D-SSIM loss. This is represented in the equation: 
\begin{align}
\mathcal{L} = \lambda \mathcal{L}1 + (1-\lambda)\mathcal{L}_{SSIM},
\end{align} where $\lambda$ is the weight parameter that balances the contribution of the $\mathcal{L}_1$ loss and the D-SSIM loss to the final objective. 

For the refinement stage, the training is supervised by all synthesized views ${I_0, I_1, ...,I_{N+M_1+M_2}}$. The optimization duration for the refinement stage generally exceeds the initialization stage. However, we still employ early stopping to prevent the formulation of incorrect Gaussian shapes.

\begin{figure*}[!ht]
    \centering
    \includegraphics[width=0.95\linewidth]{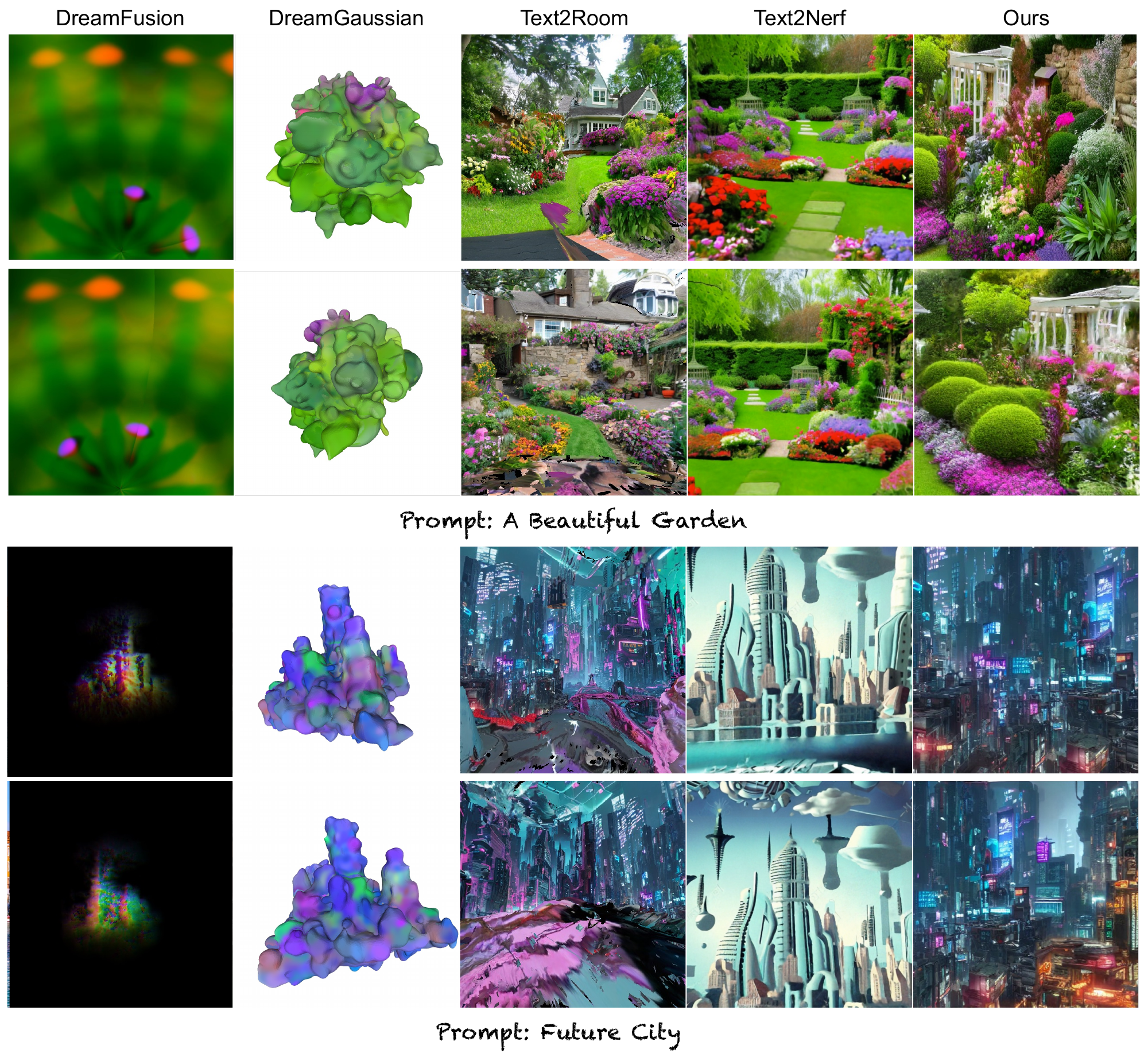}
    \caption{%
        \textbf{Qualitative comparison with baselines} including DreamFusion~\cite{poole2022dreamfusion}, DreamGaussian~\cite{tang2023dreamgaussian}, Text2Room~\cite{hoellein2023text2room}, Text2Nerf~\cite{zhang2023text2nerf} on the generation quality. We highly recommend that readers view the accompanying videos for a more thorough comparison.
    }
    \label{fig:qualitative}
\end{figure*}

\begin{figure*}[!ht]
    \centering
    \includegraphics[width=0.95\linewidth]{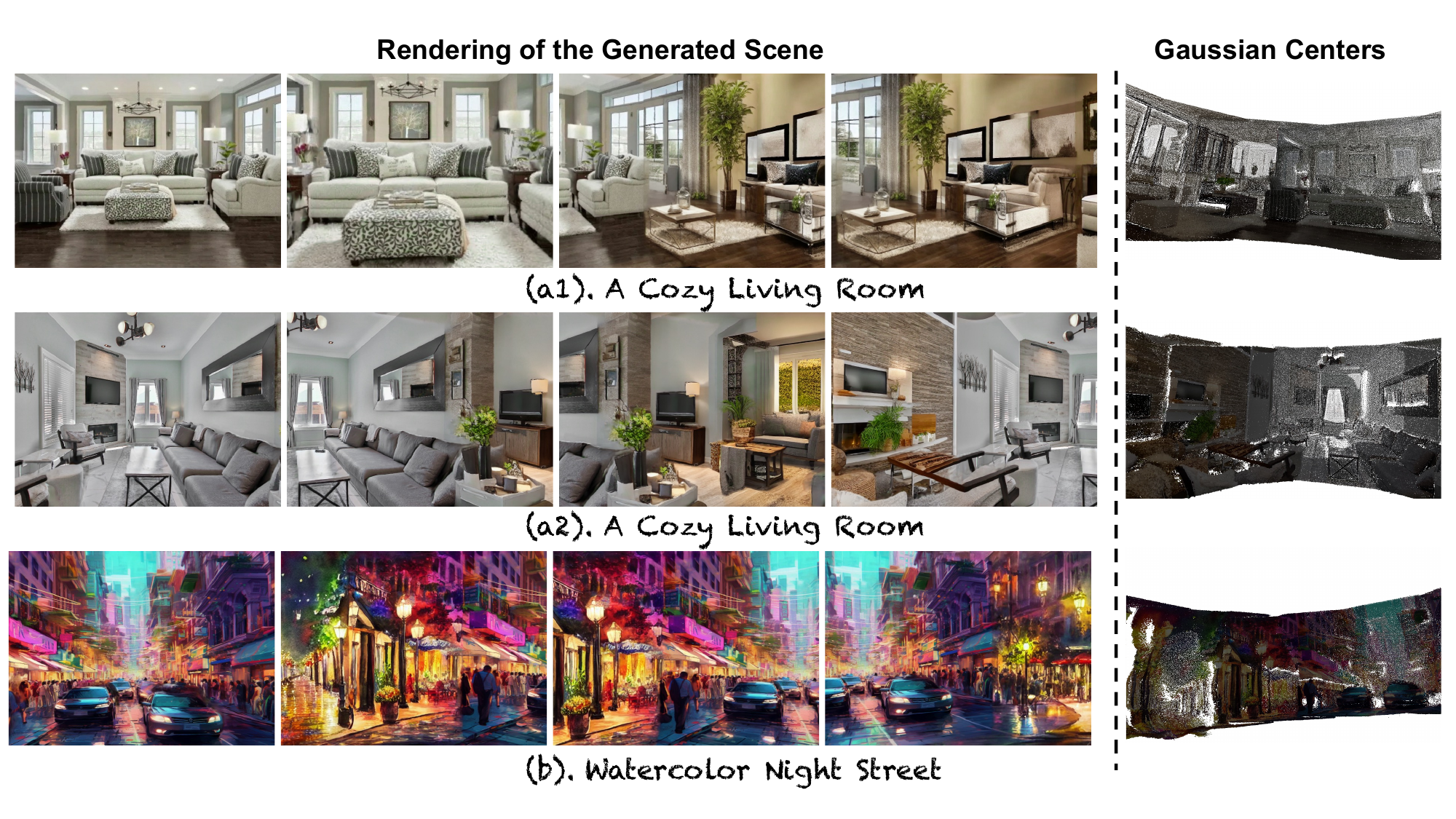}
    \caption{%
        \textbf{Diverse Output Generation:} Our pipeline is capable of synthesizing a variety of 3D scenes using the same prompts. We also demonstrate its ability to generate stylized scenes.
    }
    \label{fig:applications}
\end{figure*}

\section{Experiments}
\label{sec:experiments}
\subsection{Implementation Details}
During the first stage, we employ ZoEDepth~\cite{bhat2023zoedepth} for depth estimation and ControlNet~\cite{zhang2023adding} for outpainting; the initial optimization of 3D Gaussians is halted early at 1,000 iterations. We establish the angle difference between anchor cameras at 25 degrees and set the number of anchors between 7 and 16, contingent on the specific generation application. For the refinement stage, we utilize ControlNet~\cite{zhang2023adding} for inpainting and DiffBIR~\cite{lin2023diffbir} for super-resolution. The refinement training is also given an early stop at 2,000 iterations. We sample between 3 to 6 cameras around each anchor and incorporate additional randomly sampled, zoomed-in views. Depending on the number of cameras sampled for refinement, it takes between 10 to 25 minutes to generate each scene on a single GPU. After the optimization process, the generated Gaussian can be rendered in real-time. 

\subsection{Comparison with Baselines}  
To assess the efficacy of our approach in the context of text-driven 3D scene generation, we compare with representative methods from three distinct categories. These include methods that employ SDS loss to distill 3D from pretrained Diffusion models (e.g., DreamFusion~\cite{poole2022dreamfusion}), methods that also use 3D Gaussians as underlying 3D representations (e.g., DreamGaussians~\cite{tang2023dreamgaussian}), and methods that utilize diffusion models to generate supervisions but employ alternate 3D representations (e.g., Text2Room~\cite{hoellein2023text2room} and Text2Nerf~\cite{zhang2023text2nerf}).

\begin{table}[h]
\centering
\begin{tabular}{c|c|c|c}
\hline
Method & BRISQUE $\downarrow$ & NIQE $\downarrow$ & CLIP $\uparrow$ \\ 
\hline
DreamFusion~\cite{poole2022dreamfusion} & 67.0 & 12.0 & 22.6 \\
DreamGaussian~\cite{tang2023dreamgaussian} & 59.1 & 11.7 & 23.2 \\
Text2Room~\cite{hoellein2023text2room} & 28.4 & 5.41 & 28.1 \\
Text2Nerf~\cite{zhang2023text2nerf} & 24.5 & 4.62  & 28.7 \\
Ours & \textbf{22.7} & \textbf{4.37} & \textbf{28.9}\\
\hline
\end{tabular}
\caption{\textbf{Quantitative Comparison with Baselines}}
\label{tab:my_label}
\end{table}

\begin{table}[h]
\centering
\begin{tabular}{c|c|c|c}
\hline
Stages & PSNR & SSIM & LPIPS \\ 
\hline
Initial  & 36.3 & 0.983 & 0.019\\
Refinement  & 35.1 & 0.982 & 0.023 \\
\hline
\end{tabular}
\caption{\textbf{Reconstruction Quality}. As our refinement stage includes more images for optimization, the metrics is slightly lower.}
\label{tab:reconstruct}
\end{table}

\noindent\textbf{Quantitative Comparison} Given the absence of ground truth references for 3D scenes generated in relation to text prompts, conventional metrics such as PSNR and LPIPS~\cite{zhang2018unreasonable} are not applicable for generation tasks. Nonetheless, we include PSNR, SSIM and LPIPS~\cite{zhang2018unreasonable} values for the reconstruction of Gaussians as in Table.~\ref{tab:reconstruct}. The metrics prove the optimized Gaussians have already achieved high fidelity of reconstruction even with the early stop strategy. To assess the quality of the generated images, we follow the settings established in Text2Nerf~\cite{zhang2023text2nerf}, which employs BRISQUE~\cite{mittal2012no} and the Natural Image Quality Evaluator (NIQE)~\cite{mittal2012making} to evaluate quality in the absence of a reference image. Additionally, we utilize the CLIP text-image similarity score~\cite{radford2021learning,hessel2021clipscore} to evaluate the degree of alignment between the rendered images and the input prompt. In both the image quality and alignment with text prompts, our method achieves the highest results. However, as these metrics may not provide a fully accurate representation, qualitative comparisons remain a crucial aspect of our evaluation process.


\noindent\textbf{Qualitative Comparison}
We present qualitative comparison results with other baseline methods. As depicted in the Fig.~\ref{fig:qualitative}, previous works like DreamFusion~\cite{poole2022dreamfusion} which primarily focus on the generation of 3D objects underperforms in the context of scene generation, leading to blurry and less realistic textures. DreamGaussian utilize the 3D Gaussians as 3D representations but also works only on object-centric generation. When compared with Text2Room~\cite{hoellein2023text2room}, our results capably handle outdoor scenes without noticeable gaps, owing to the high capacity of the Gaussians. Against Text2Nerf~\cite{zhang2023text2nerf}, our results achieve superior rendering quality, leading to more realistic scene generation. Note that as Text2Nerf has not been open-sourced yet, we directly use the released results with the same text prompts. However, image comparison does not capture 3D consistency. Therefore, we strongly recommend that readers refer to the supplementary videos for a more comprehensive comparison.

\begin{figure}[!ht]
    \centering
    \includegraphics[width=0.97\linewidth]{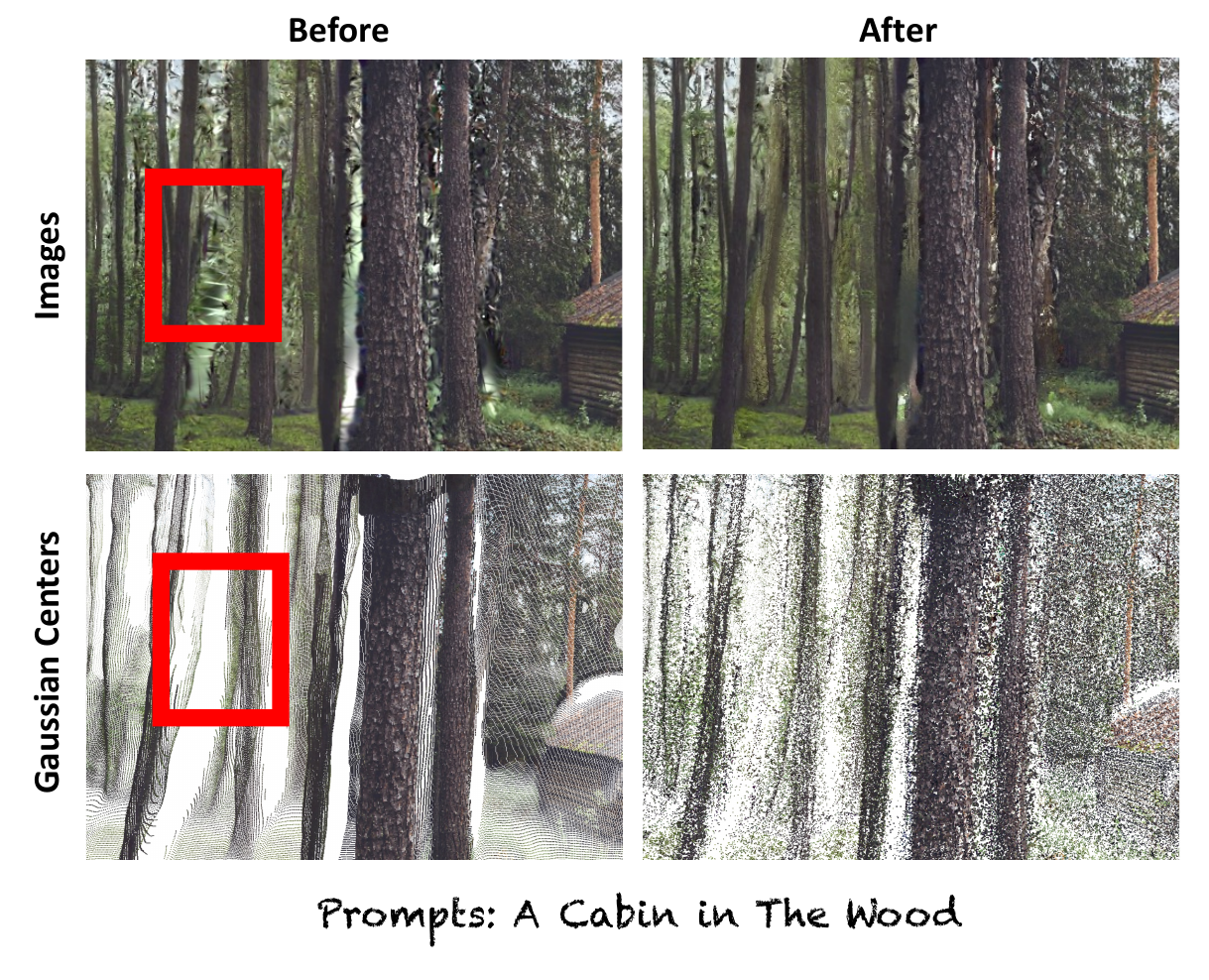}
    \caption{%
        \textbf{Ablation study} on the refinement stage, which shows that the refinement helps filling in the missing regions and reducing the noisy appearance. After the refinement, the center of Gaussians effectively covering the original missing parts. 
    }
    \label{fig:ablation}  
\end{figure}

\begin{figure}[!ht]
    \centering
    \includegraphics[width=0.97\linewidth]{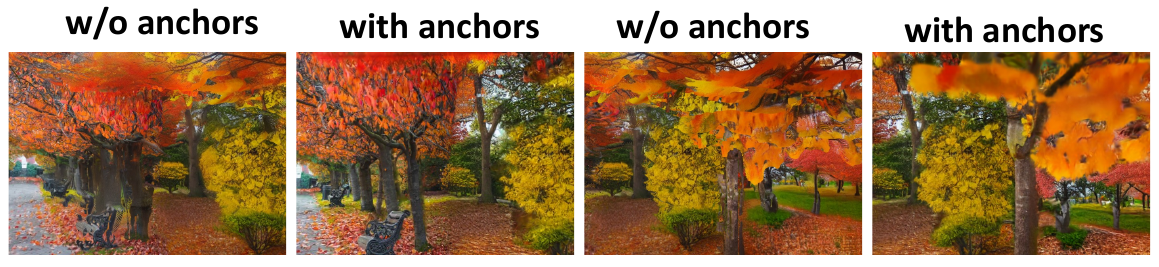}
    \caption{%
        \textbf{Ablation study} on the initialization. Without using the anchor cameras, the generated images contain obvious gap. 
    }
    \label{fig:ablation2}  
\end{figure}

\subsection{Generation Capability}
We showcase the high generation capacity and robustness of the proposed pipeline in the following three areas:
\noindent\textbf{Diversity:} As depicted in the accompanying figures, our pipeline can generate a diverse array of photorealistic 3D scenes using the same text prompts as in Fig.~\ref{fig:applications}.
\noindent\textbf{Stylized and Imaginary Scenes:} As demonstrated in the figure, our approach is also capable of generating stylized and imaginary scenes in Fig.~\ref{fig:applications}.
\noindent\textbf{360 VR View:} Our method can further generate 360-degree views, suitable for VR scenes. The results are attached in the supplement. 

\subsection{Ablation Study}
We conduct ablation studies to examine the design elements of our pipeline. Primarily, we are interested in investigating how refinement contributes to the enhancement of rendering quality and how the process of point initialization impacts the results. More ablations about early stop and removing stretched points are attached in the supplement.

\noindent\textbf{Ablation on Refinement} We examine the effectiveness of the refinement process as illustrated in the corresponding figures. In the absence of inpainting, it is observable that missing parts are filled with optimized Gaussians. Without the restoration, the zoomed-in views are populated with Gaussian shapes, resulting in a noisy appearance. Following the refinement process, the rendered image exhibits enhanced cleanliness and clarity.

\noindent\textbf{Ablation on Gaussian Initialization} We explore the significance of 3D Gaussians initialization as in Fig.~\ref{fig:ablation2}.  If we initialize the Gaussian cloud directly using all views, as opposed to solely using the anchor views, the point cloud becomes disorganized, which subsequently impacts the quality of the generated Gaussians. The randomness of the viewing angle decrease the depth estimation accuracy especially for close views. 


\section{Limitation and Conclusion}
\label{sec:conclusion}
\begin{figure}[t]
    \centering
    \includegraphics[width=0.97\linewidth]{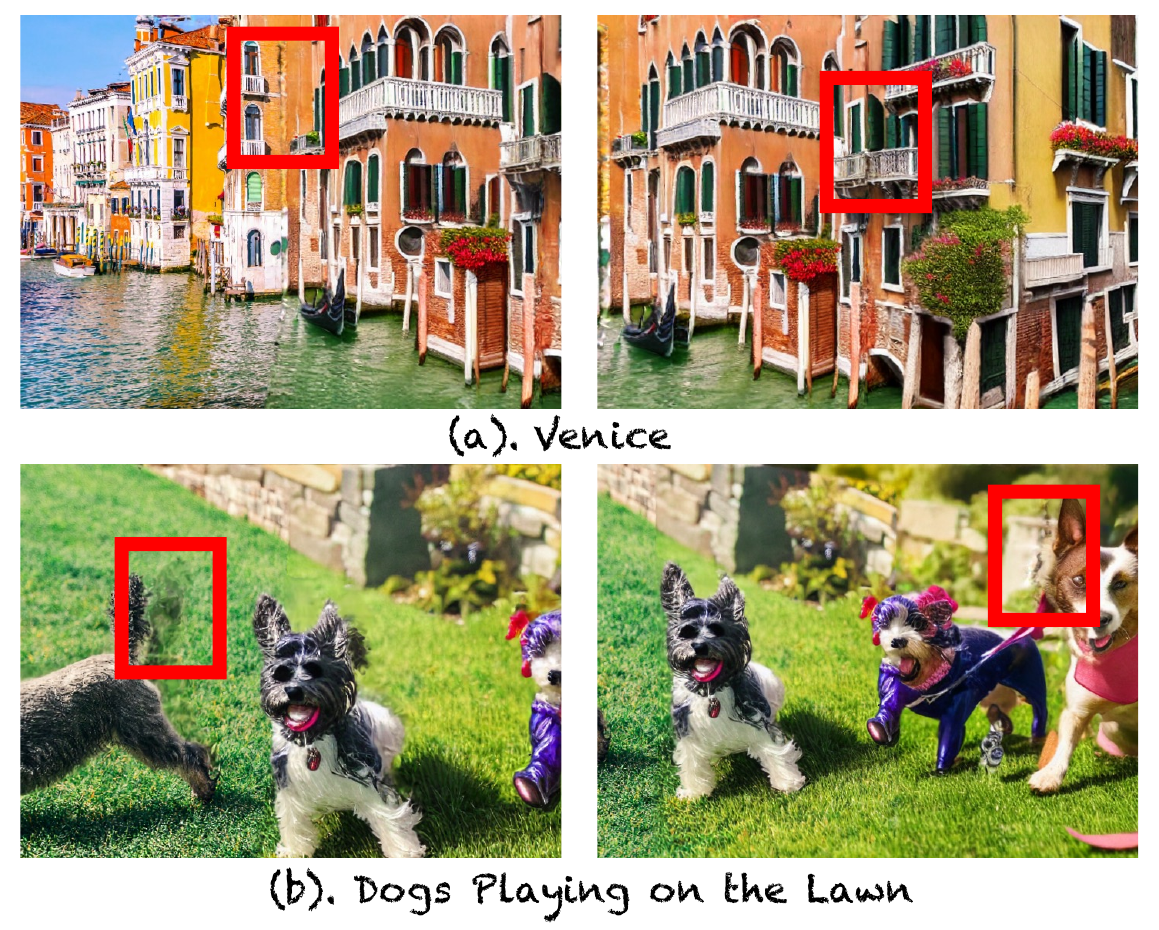}
    \caption{%
        \textbf{Failure cases}. In (a), we demonstrate a scenario where an erroneous alignment results in noticeable gaps. In (b), we highlight an instance where the use of inpainting leads to undesirable ghosting effects.
    }
    \label{fig:failure cases}  
\end{figure}


In this study, we delve into the generation of immersive 3D scenes using 3D Gaussians. Our approach has yielded promising results, yet it has also revealed several challenges that necessitate further exploration. As depicted in Fig.~\ref{fig:failure cases}, a key constraint of our method is the initial construction phase of the Gaussian cloud, which is heavily dependent on monocular depth estimation. Although depth alignment and subsequent Gaussian training can alleviate misalignment issues, significant estimation errors may still yield noticeable visual artifacts. Furthermore, the introduction of new objects during the refining stage can cause ghosting effects.

Nevertheless, despite these challenges, our proposed pipeline generally delivers high-fidelity 3D scene generation with 3D consistency, opening the door to various 3D applications. Future research is expected to enhance the results of inpainting 3D Gaussians, especially in cases where only 2D inputs are available.
{
    \small
    \bibliographystyle{ieeenat_fullname}
    \bibliography{main}
}

\section*{Appendix}
\setcounter{section}{0}
\appendix
The supplementary materials are organized as follows: Initially, we delve into the specifics of our implementation, detailing the training settings and the construction of the pointcloud. Following this, we execute a more comprehensive analysis of several issues identified in the paper. Finally, we present additional results, featuring a diverse range of text prompts.

\section{Implementation Details}

\subsection{Model Details} 
For Gaussian training, we initialize the opacity at 0.5 and set the initial shape as a sphere. We've observed that the training of 3D is significant. The image width and height are configured at 704 and 512, respectively. Adhering to the original settings in Gaussian splatting, the learning rate of opacity is established at 0.05, scaling at 0.005, and rotation at 0.001. The split and clone process is triggered every 100 iterations.

During the inpainting and outpainting process, in the initial stage, we set the mask dilation pixels to 14 and adjust it to 5 for the refinement stage. For ControlNet~\cite{zhang2023adding}, we set the scale at 9 and the control strength at 1. The guess mode is deactivated and the generation step is set to 20.

Following the configuration in ZoEDepth~\cite{bhat2023zoedepth}, where the field of view is set to an angle of 55 degrees, we adopt these settings for the intrinsics of our synthetic cameras.

\begin{figure*}[t]
    \centering
    \includegraphics[width=0.95\linewidth]{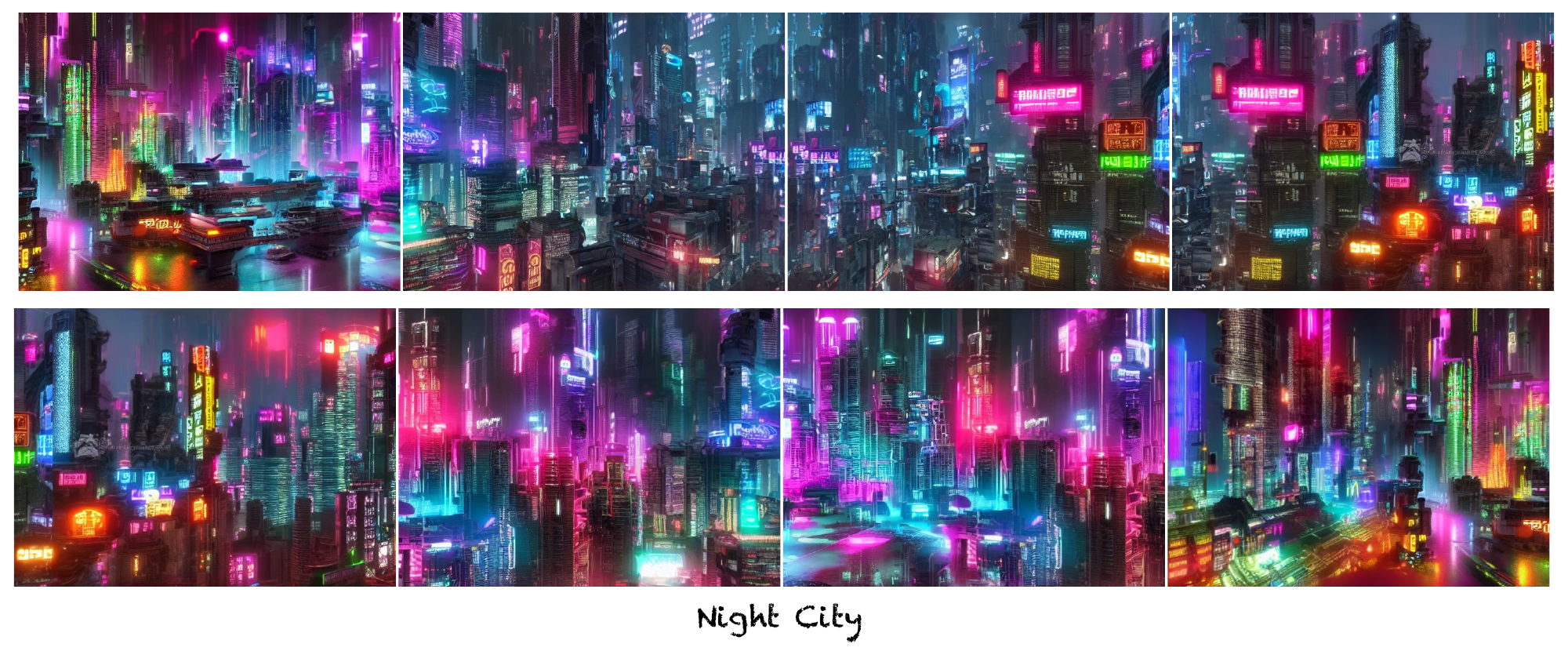}
    \caption{%
        \textbf{360 Scenes.}.
    }
    \label{fig:360}
    \vspace{-7pt}
\end{figure*}

\subsection{Depth Alignment}  

In this section, we delve deeper into the process of depth alignment, an integral part of our methodology. The final step in our procedure involves a crucial update where the existing points are updated with newly generated ones. This step requires attention as it involves aligning the newly estimated depth map, represented as $D$, with the pre-existing components of the model.

 A crucial aspect of this process is the minimization of error in the overlapping regions. These areas, where the new points intersect with the existing ones, can be particularly prone to discrepancies. By focusing on these overlapping parts, we aim to optimize the alignment and, by extension, the overall quality of the resulting model. To achieve this alignment, we estimate two key parameters: the global scale $s$, and the global depth shift $b$. The global scale $s$ refers to the factor by which we adjust the depth values of the new points to match the scale of the existing model. On the other hand, the global shift $b$ corresponds to the uniform adjustment required to align the depth values of the new points with those of the existing ones. To determine the optimal values of $s$ and $b$ that minimize the error in the overlapping regions, we frame this as a least squares problem.  The method involves finding the values of $s$ and $b$ that minimize the sum of the squares of the differences between the estimated and actual depth values in the overlapping regions.

We compute $s$ and $b$ using the following least squares formulas:

\begin{align*}
s &= \frac{n \sum{D D_0} - \sum{D} \sum{D_0}}{n \sum{D^2} - (\sum{D})^2} \\
b &= \frac{\sum{D_0} - s \sum{D}}{n},
\end{align*}

where $n$ is the total number of the points, and $D_0$ is the original depth. The depth after the alignments becomes:  

\begin{align*}
D' = s D + b
\end{align*}

\section{Analysis}
We carry out additional experiments to delve into three aspects: an ablation study examining the impact of removing stretched points during initialization, an investigation into the position of newly added Gaussians via the split-and-clone method, and a comparison study to assess the effect of applying early stopping. 

\subsection{Removing Stretched Points} 
\begin{figure}[t]
    \centering
    \includegraphics[width=0.97\linewidth]{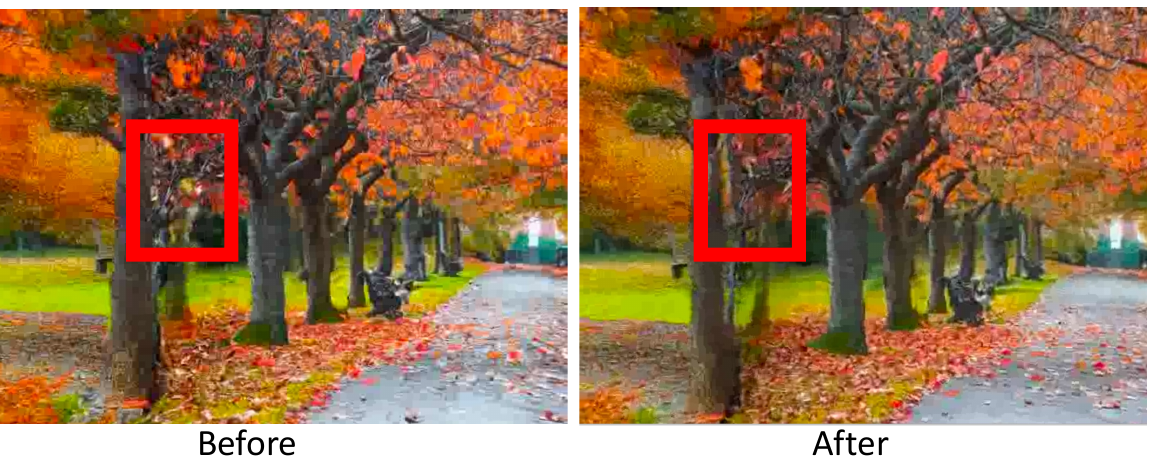}
    \caption{%
        \textbf{Comparison on Removing Stretched Points.} 
    }
    \label{fig:stretched}  
\end{figure}

During the initialization phase, we discard the stretched points because their incorrect initial positions could potentially degrade the quality of the inpainting results. Fig.~\ref{fig:stretched} demonstrates that the removal of these points results in a clearer inpainted region.

\subsection{Analysis on Split-and-Clone} 
\begin{figure}[t]
    \centering
    \includegraphics[width=0.97\linewidth]{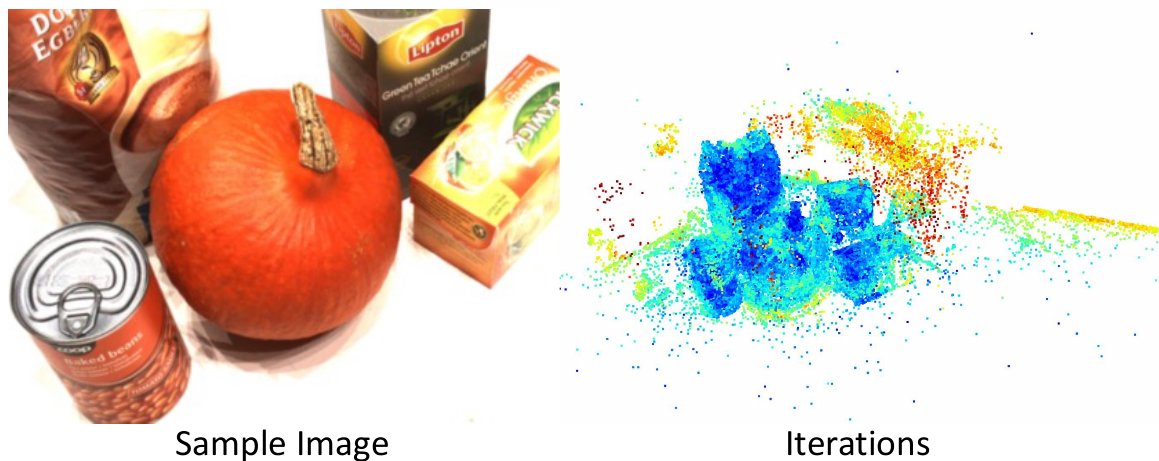}
    \caption{%
        \textbf{Analysis on the Newly Splitted Points.}. 
    }
    \label{fig:split}  
\end{figure}

We've observed a notable pattern with the split-and-clone operation in Gaussian Splatting. Each Gaussian is tracked with an additional parameter that records the iteration of splitting. If a Gaussian splits at the $N$th iteration, this parameter is recorded as $N+1$. As depicted in Fig.~\ref{fig:split}, the blue color represents initial points. The redder a point becomes, the higher its iteration number. It can be seen that the points split progressively from near to far. With this pattern in place, assuming the initial points are correct, newly added points will be positioned around the existing surface. This characteristic is advantageous during training when only sparse 2D supervision is available.

\subsection{Comparison on Early Stop} 
\begin{figure}[t]
    \centering
    \includegraphics[width=0.97\linewidth]{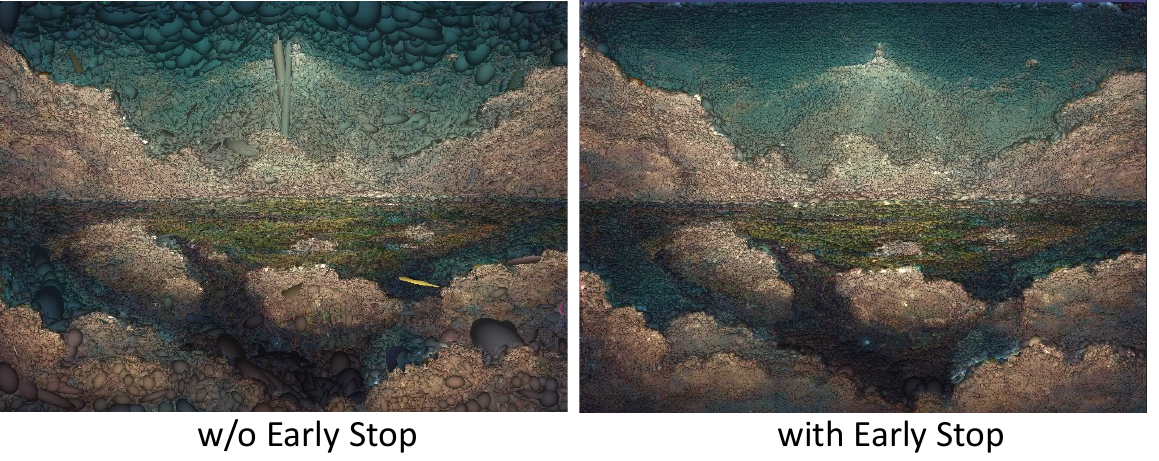}
    \caption{%
        \textbf{Analysis on Early Stop.}. 
    }
    \label{fig:early}  
\end{figure}

We present a comparison of results with and without the implementation of early stopping. As illustrated in Fig.~\ref{fig:early}, the accuracy of the Gaussian shapes degrades when early stopping is not applied.

\section{More Results}
In this section, we include additional results generated using our proposed Text2Immersion pipeline. We also provide videos that better illustrate the 3D consistency. We strongly encourage readers to examine these materials.  

\begin{figure*}[t]
    \centering
    \includegraphics[width=0.95\linewidth]{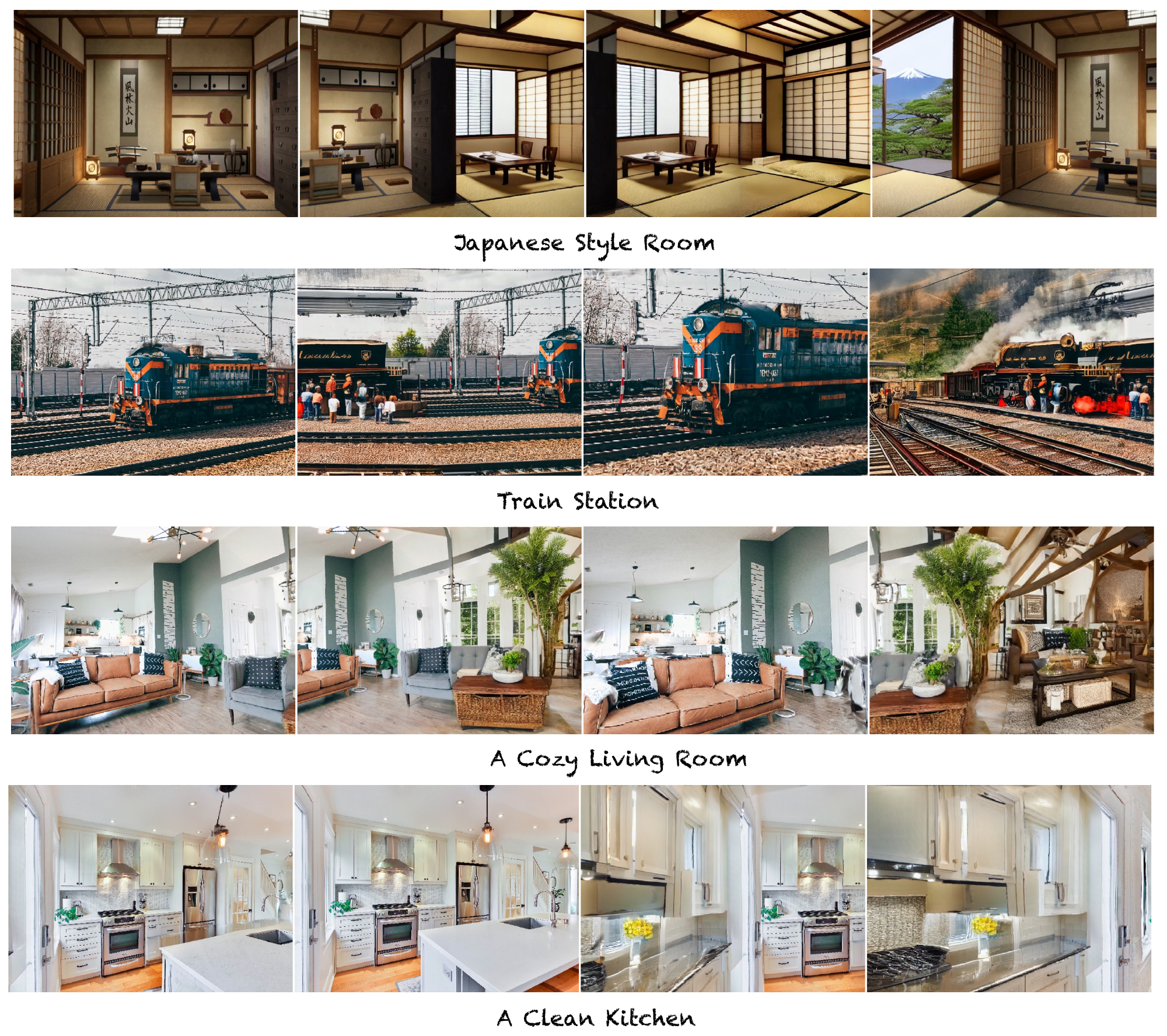}
    \caption{%
        \textbf{More results.}.
    }
    \label{fig:more_results}
    \vspace{-7pt}
\end{figure*}

\subsection{More Results with Diverse Text Prompt}
We provide more results with diverse text prompt in Fig.~\ref{fig:more_results}.
\subsection{360 Scenes}
We also show the 360 scenes Fig.~\ref{fig:360}.

\end{document}